\documentclass[10pt,twocolumn,letterpaper]{article}

\usepackage[pagenumbers]{cvpr}

\usepackage{float}
\usepackage{subcaption}
\usepackage{caption}
\usepackage{array}

\definecolor{cvprblue}{rgb}{0.21,0.49,0.74}
\usepackage[breaklinks,colorlinks,allcolors=cvprblue]{hyperref}

\title{Event-Based Dynamic Lip Region Estimation During Speech}

\author{Hamze Hammami\\
Independent Researcher\\
Dubai, UAE\\
{\tt\small hamzehousamhammami@gmail.com}
\and
Omar Khaled Elsayed Mohamed Eldardeer\\
Italian Institute of Technology (CONTACT Unit)\\
Genova, Italy\\
{\tt\small omar.eldardeer@iit.it}
\and
Francesco Rea\\
Italian Institute of Technology (CONTACT Unit)\\
Genova, Italy\\
{\tt\small francesco.rea@iit.it}
}

\begin{document}
\maketitle
\begin{abstract}
Event cameras offer high temporal resolution and low power consumption, making them promising for speech-related applications. These include automatic lip reading, audiovisual speech analysis, and human-robot interaction, particularly in challenging environments characterized by low illumination or high dynamic range. Event-based lip reading, speaker recognition, and voice activity detection, however, mostly operate on a cropped region of interest (ROI), by hand or by the dataset, and no event-based method supplies that ROI as a spatial region; existing event-based face analysis gives the mouth as landmarks, which presents a mismatch between application and localization. The present work predicts a dense heatmap over the lip region from the full event "frame". Events are encoded as a polarity-separated voxel grid and processed at every location by a two-pathway continuous-time recurrent network (CT-RNN) with learned, log-space bounded time constants acting on the real elapsed time of each bin: a stateless fast pathway and a stateful slow pathway, utilised for refining and anchoring the prediction. The model reaches 69.5\% on lips on four unseen speakers and switches target only 2.1 times per second, against 8.9 and 11.0 for convolutional baselines with $5.7\times$ more parameters. Adding a spatial encoder improves precision but fails on a new face scale. On a real event sensor, trained only on simulated events, the proposed model reaches 58.0\% on lips against 43.2\% for the CNN, with fewer than half its target switches.
\end{abstract}
\section{Introduction}
\label{sec:intro}

Neuromorphic event cameras report per-pixel brightness changes asynchronously, producing a stream of timestamped events with a microsecond temporal resolution, high dynamic range, low latency, and low power consumption~\cite{gallego2020event}. This makes them well-suited to rapid facial dynamics such as blinks and gaze, which demand a temporal precision beyond what fixed-rate sensors provide. These advantages do not transfer for free: a survey of neuromorphic face analysis~\cite{becattini2024neuromorphic} shows that detectors and landmark estimators built for RGB frames degrade sharply on event streams, producing low-confidence or degenerate predictions even on a clearly visible face, and that the field still lacks standardized, event-native detectors. As a result, event-based applications that need the face or mouth fall back on frame-based detectors or manual annotation. Lip localization matters most for speech. Visual speech carries predictive information for hearing: mouth movements precede the voice by 100 to 300\,ms~\cite{chandrasekaran2009natural}, and seeing them speeds up the cortical processing of the auditory signal~\cite{van2005visual}. Listeners fuse the two streams within a temporal window of about 200\,ms~\cite{van2007temporal}, which a 25 to 30\,fps camera covers with only five or six frames, so fast articulatory movements such as lip closures and releases fall between frames. Event cameras record these movements at a much higher temporal resolution and with less redundant information~\cite{tan2022multi}, which motivates event-based lip reading and speech prediction, including applications for human--robot interaction. Event-based lip reading, speaker recognition, and voice activity detection all depend on the mouth region of interest (ROI), which, in the current applications, is either cropped by hand or localized from the events only as a point or a set of landmarks.

\subsection{Related Work}
Face detection with event cameras was first shown on event-derived patches~\cite{barua2016direct}; \cite{lenz2020event} correlated ON and OFF activity against a temporal model of an eye blink, giving the first purely event-based face detector, and \cite{ryan2021real} learned face and eye tracking on simulated events, again relying on blinks.

Event-based lip reading recognizes spoken words from mouth motion. \cite{tan2022multi} introduced DVS-Lip, the first event-based lip-reading dataset, released with the lip region already extracted to a $128 \times 128$ window, and classified words by fusing a low- and a high-frame-rate branch. \cite{bulzomi2023end} proposed an end-to-end spiking network for the task, aimed at the low latency of embedded neuromorphic processing, and \cite{dampfhoffer2024neuromorphic} extended spiking lip reading with signed spiking gated recurrent units.
\cite{yoo2024rn} reached competitive accuracy on DVS-Lip by attaching a reservoir node with fixed dynamics to every pixel, a per-pixel temporal encoding without a learned time constant. All four take the pre-extracted lip window as their input, so the crop is performed by the dataset rather than by the method, and none estimates where the lips are within a wider view. Beyond word recognition, \cite{yao2026neurolip} builds a speaker recognition system from lip motion recorded with an event camera;  the lip region of each recording is annotated by hand and every sample is cropped to a $200 \times 160$ window that keeps only the events inside it. \cite{kanamaru2023isolated} does localize the lips before recognition, but with frame-based face and facial-feature detectors applied to a co-recorded conventional camera and to intensity images reconstructed from the events, rather than from the event stream itself.

\begin{table*}[t]
\centering
\footnotesize
\caption{Event-based prior work, compared by what each method outputs for the mouth and which crop it needs as input.}
\label{tab:lip_prior_work}
{\renewcommand{\arraystretch}{1.15}
\setlength{\tabcolsep}{4pt}
\begin{tabular}{@{}>{\raggedright\arraybackslash}p{1.2cm}>{\raggedright\arraybackslash}p{2.9cm}>{\raggedright\arraybackslash}p{3.4cm}>{\raggedright\arraybackslash}p{3.3cm}>{\raggedright\arraybackslash}p{2.7cm}>{\raggedright\arraybackslash}p{2.4cm}@{}}
\toprule
\textbf{Method} & \textbf{Objective} & \textbf{Event representation} & \textbf{Model} & \textbf{Mouth output} & \textbf{Input crop} \\
\midrule
\cite{tan2022multi} & Lip reading (word classification) & Event frames at two frame rates & Two branches, message flow fusion & None, window given & Mouth, by dataset ($128 \times 128$) \\
\cite{yao2026neurolip} & Speaker recognition & Polarity-separated voxels, learned temporal allocation & Modified ResNet34 & None, window given & Mouth, by hand ($200 \times 160$) \\
\cite{savran2018energy} & Voice activity detection & Hand-crafted Gabor filter & MAP estimate over a coarse grid & Lip centre (one point) & None, optional face prior \\
\cite{kielty2026locally} & Face and five-landmark detection & Locally adaptive decay surface & MobileNetV3, PIPNet head & Mouth corners (2 of 5 points) & Face, by detector \\
\cite{kielty2025event} & 98-landmark and blink detection & Histogram or time surface, summed polarity & MobileNetV2, ConvLSTM, PIPNet head & Mouth contour (about 20 of 98 points) & Face, by detector \\
\midrule
Proposed & Lip region localization & Polarity-separated voxel grid ($K$ bins) & Two-pathway CT-RNN with persistent slow memory & Dense lip region (heatmap) & None \\
\bottomrule
\end{tabular}}
\end{table*}

For lip localization, \cite{savran2018energy} estimates the lip centre as a single point, only as the front end to a voice activity detection (VAD) classifier.  \cite{savran2023fully} continues this VAD line but drops lip localization altogether, taking the mouth region directly from manual landmarks and collapsing its spatial structure into a temporal signal; joint spatial localization is explicitly set aside there as a harder problem. More recently, \cite{kielty2026locally} places two of its five facial landmarks at the mouth corners while pursuing face detection and general landmark localization, and \cite{kielty2025event} predicts a ninety-eight-point face, with around twenty points along the mouth contour, for landmark and blink detection in driver monitoring. In both, the mouth points are read off a whole-face model after the face is cropped by a separate detector. 

\subsection{Problem Statement}

Taken together, existing work leaves three gaps:
\begin{itemize}
\item \textbf{Task and output:} Applications of event-based lip analysis consume a mouth region: lip reading runs on data already restricted to the mouth~\cite{tan2022multi}, speaker recognition on a hand-cropped lip window~\cite{yao2026neurolip}, and voice activity detection on a mouth region taken from manual landmarks~\cite{savran2023fully}. The methods that do locate the mouth return a single point~\cite{savran2018energy} or a few landmarks from a whole-face model that itself needs a face crop~\cite{kielty2025event}, so a region must still be rebuilt before any of these applications can run.

\item \textbf{Representation:} HOTS-style time surfaces~\cite{lagorce2017hots} and locally adaptive decay surfaces~\cite{kielty2026locally} collapse a window of events into a single 2D surface in which each pixel holds a decayed summary of its recent activity, so the order of events within the window is lost. Event histograms keep polarity but discard timing in the same way. Voxel grids~\cite{zhu2019unsupervised} keep the timing in bins, but in their standard form sum signed polarity into one channel. Learned grid representations~\cite{gehrig2019end} can keep both, but at an added computational cost. Polarity is handled inconsistently across these representations: time-surface methods keep ON and OFF on separate surfaces~\cite{lagorce2017hots}, while ordinal surfaces such as EROS~\cite{gava2022puck} drop polarity altogether; where polarity is merged into one channel, opposite polarities can cancel at the lip boundary and erase the signed edge structure that marks the moving contour.

\item \textbf{Temporal stability:} Some of the methods in Table~\ref{tab:lip_prior_work} carry temporal state, through recurrent layers~\cite{kielty2025event} or decaying surfaces~\cite{kielty2026locally}, yet methods that work on localization are evaluated on per-frame accuracy, such as landmark error or the share of frames localized within a tolerance~\cite{savran2018energy,kielty2025event,kielty2026locally}. A lip ROI should remain on the mouth from one window to the next, a property that has not yet been measured for this task.
\end{itemize}

This work proposes a solution to these gaps: a model that outputs the lip region from the event stream, and an evaluation of its temporal stability. To the authors' knowledge, no event-based method treats the lip ROI as its primary objective: the closest estimates only the lip centre for voice activity detection~\cite{savran2018energy}, and landmark methods read the mouth off a whole-face model after a face crop~\cite{kielty2025event,kielty2026locally}. The proposed model goes beyond these by predicting a dense lip heatmap from the full, uncropped event frame, as a front end for lip analysis. A two-pathway continuous-time recurrent network (CT-RNN)~\cite{funahashi1993approximation}, using the time-aware update of liquid time-constant networks~\cite{hasani2021liquid}, reads the events at every location: a stateless fast pathway follows lip motion, and a stateful slow pathway keeps a memory that anchors the prediction.

The contributions of this work are:
\begin{itemize}
\item Event-based lip localization formulated as a front end for lip analysis, predicting a dense lip region from the full uncropped event frame.

\item A two-pathway CT-RNN applied per spatial location over polarity-separated bins, with learned time constants acting on real elapsed time, a stateless fast pathway and a stateful slow pathway, trained on a pooled RGB teacher whose uncertainty the network also predicts.

\item An evaluation on unseen speakers against frame-wise convolutional baselines that reports temporal stability alongside accuracy, and an analysis of how spatial context trades accuracy against robustness to new speakers.

\item A transfer to real event sensors in which, with all models trained on simulated data, the proposed model outperforms the CNN in both on-lips rate and stability.
\end{itemize}

The paper describes the data, the proposed model and the baselines, evaluates accuracy and temporal stability on unseen speakers and a real event sensor, and discusses the results against the gaps above.
\section{Data}
\label{sec:data}
Training uses one simulated set; evaluation adds sequences of speaking faces selected from two real event-camera datasets. The simulated set contains 12 speakers (8 male, 4 female) recorded with a conventional camera at 30 frames per second while reading teleprompter prompts at a natural pace, with no face or mouth crop. Speakers S2--S7 are used for training, S8--S9 for validation, and S10--S13 are held out as unseen test speakers; S1 is a mock user and is not used. The real sets are Faces in Event Streams (FES)~\cite{s24051409}, which provides face boxes and five facial landmarks including the mouth corners, and FACEMORPHIC~\cite{becattini2024neuromorphic2}, which records RGB video alongside the event stream, although the two are not spatially aligned.

\textbf{Event simulation.} Each clip is upsampled from 30\,Hz to 320\,Hz with the frame interpolation of VID2E~\cite{gehrig2020video} and converted to events with ESIM~\cite{rebecq2018esim} at a contrast threshold of $0.2$; one event transition spans two consecutive interpolated frames, 6.25\,ms. Coordinates and timestamps are normalised to $[0,1]$; windows above 20k events (corrupt interpolations) are dropped and the rest capped at 10k.

\textbf{Supervision target (teacher).} Three face parsers, FaRL~\cite{zheng2022general}, a SegFormer~\cite{xie2021segformer} face parser and MediaPipe FaceMesh~\cite{lugaresi2019mediapipe}, are run on a crop of 1.6 times the RetinaFace~\cite{9157330} face box in each RGB frame. Each gives a probability for the whole mouth region (both lips and the inner mouth), so an open mouth remains a single region. The probabilities are averaged in logit space into $p$, with per-pixel confidence $c_t = 1 - H_b(p)$, $H_b$ the binary entropy, which is lowest along the mouth boundary. Each event takes the teacher value at its pixel; these values are averaged per cell of a $64 \times 64$ grid, blurred ($\sigma = 0.9$ cells) and normalised to a peak of one to form the target $T$, and $c_t$ is rasterised the same way without blur. The teachers are used only for training.
\section{Method}
\label{sec:method}
\begin{figure*}[!t]
    \centering
    \includegraphics[width=0.8\textwidth]{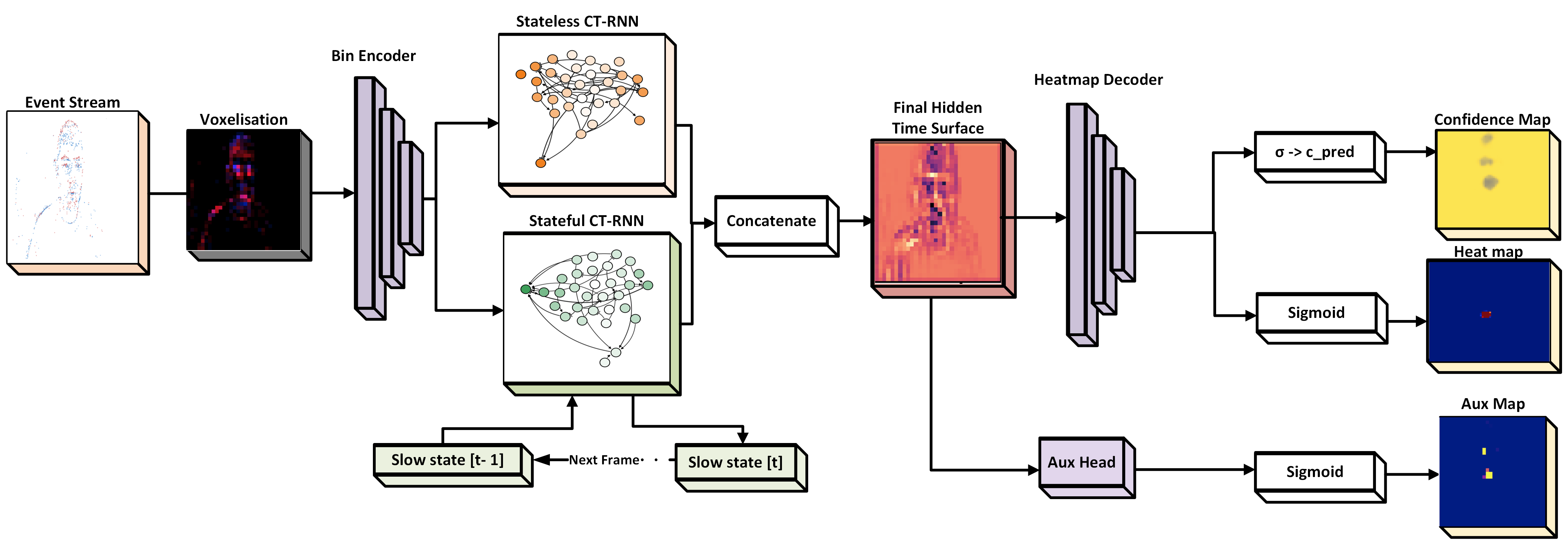}
    \caption{Overview of the proposed event-based lip-localization architecture.}
    \label{fig:lip-detector}
\end{figure*}

Figure~\ref{fig:lip-detector} shows an overview of the proposed architecture.

\subsection{Polarity-Separated Voxel Grid}
The input to the network is one 6.25\,ms event window, with shape $[x, y, t, p]$.
The events are placed into a voxel grid of $K = 16$ temporal bins on a
$G \times G$ spatial grid with $G = 32$, following the voxel-grid construction
of~\cite{zhu2019unsupervised}. ON and OFF events are written to two separate channels, and each cell holds the number of events that landed in it, passed through a $\log(1+x)$ compression
that brings sparse and dense windows onto a comparable scale while keeping their
relative ordering. This forms a separated voxel grid of shape $[K \times 2 \times G
\times G]$.

\begin{figure}[ht]
    \centering
    \includegraphics[width=0.9\linewidth]{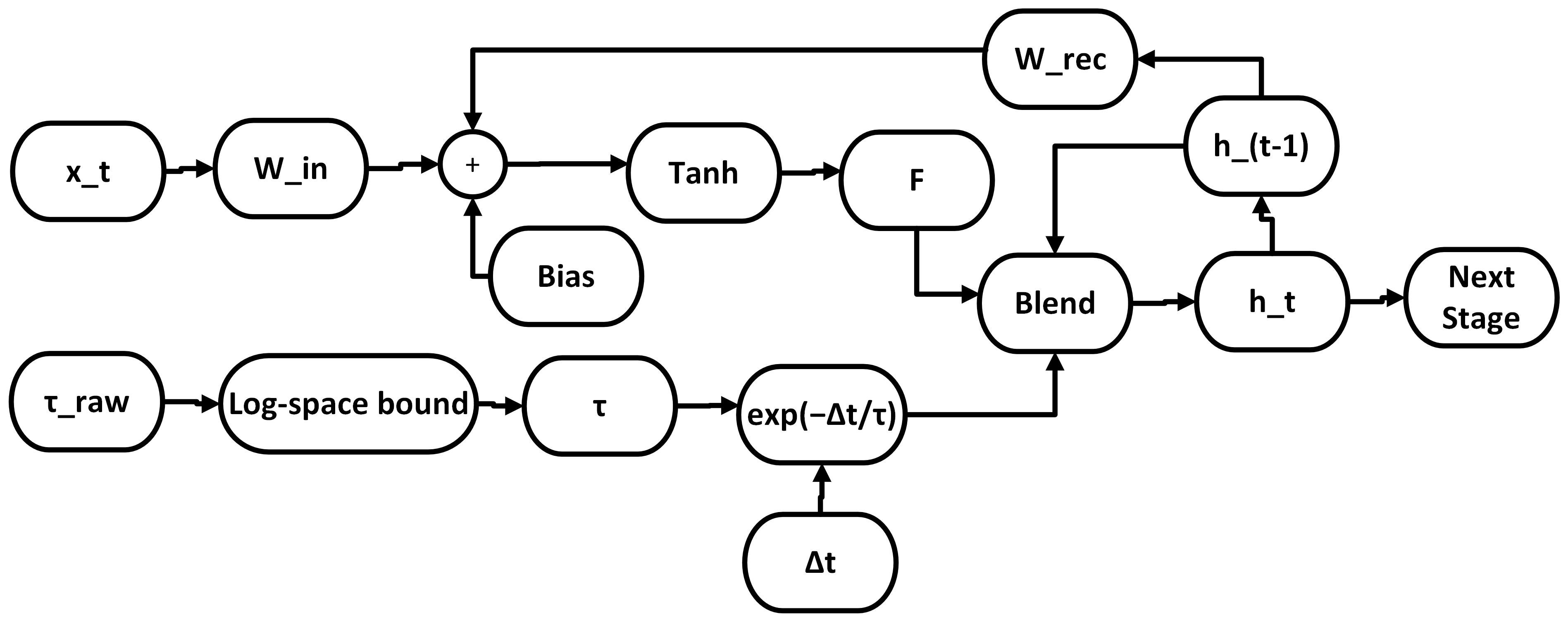}
    \caption{Structure of the continuous-time cell.}
    \label{fig:ltc-cell}
\end{figure}

Each bin is passed through a two-layer convolution encoder with $3 \times 3$
filters and 16 channels.

\subsection{Two-Timescale Continuous-Time Memory}
\label{sec:ltc}
A temporal stage is applied independently at every spatial location, across the $K$ bins, with weights shared over locations in the manner of a $1 \times 1$ convolution applied recurrently. At each location the encoded bins are read in order by a continuous-time recurrent cell (CT-RNN)~\cite{funahashi1993approximation} in rate form, $\tau\,\dot h = -h + f$, where each unit carries its own learned time constant $\tau$. Rather than an Euler update, the cell uses an exponential step that holds $f$ over the real bin duration, Equation~\ref{eq:cte}:
\begin{equation}
h \leftarrow f + (h - f)\,\exp(-\Delta t / \tau),
\label{eq:cte}
\end{equation}
where $\Delta t$ is the duration of a bin and $f$ is the candidate activation,
\begin{equation}
f = \tanh\!\big(W_{\mathrm{in}}\,x_t + W_{\mathrm{rec}}\,h + b\big),
\label{eq:cand}
\end{equation}
computed from the encoded bin $x_t$ and the previous state $h$. The update follows the closed-form, time-aware update of liquid time-constant (LTC) networks~\cite{hasani2021liquid}. Initially an LTC was chosen, but as the time constant was kept constant per unit in two timescales instead of a liquid state, it has become a CT-RNN. This lets the update act on the real duration of a bin, so the temporal behaviour depends on elapsed time; the constant $\tau$ keeps each pathway in its timescale, and the state stays bounded, at a small parameter count. Each time constant is learned through a raw parameter $\tau_{\mathrm{raw}}$ and bounded to $[\tau_{\min}, \tau_{\max}]$ by a sigmoid in log space, Equation~\ref{eq:tau}:
\begin{equation}
\tau = \exp\!\big(\log\tau_{\min} + (\log\tau_{\max} - \log\tau_{\min})\,\sigma(\tau_{\mathrm{raw}})\big).
\label{eq:tau}
\end{equation}
Bounding $\tau$ in log space keeps a gradient at every value, so no unit is frozen at a bound as with a hard clamp, and lets one optimiser step change $\tau$ by a ratio, so units from milliseconds to seconds are trained at a comparable rate.

Two pathways of 32 units run in parallel and their states are concatenated. Each $\tau$ is a learned constant.

\begin{itemize}
\item \textbf{Fast pathway:} $\tau$ constrained to $0.001$--$0.15\,\mathrm{s}$, intended for short edge bursts. It is reset at the start of every window, making it stateless.

\item \textbf{Slow pathway:} $\tau$ constrained to $0.02$--$2.0\,\mathrm{s}$, integrating sustained movement. Its state is not reset: it carries over from one window to the next, so it spans multiple windows.
\end{itemize}

A silent state decays towards its no-input value at the rate set by its own $\tau$, so the state acts as a learned time surface whose decay is learned per unit rather than fixed by hand. The concatenated state after the last bin gives a hidden map over the spatial grid. Figure~\ref{fig:ltc-cell} shows the structure of the cell.

\subsection{Heatmap, Uncertainty and Auxiliary Map}
The hidden map is passed to a convolutional decoder, a short stack of
$3 \times 3$ convolutions with group normalisation and GELU activations,
followed by a $1 \times 1$ convolution and bilinear upsampling to a
$64 \times 64$ output. Group normalisation is used in place of batch
normalisation because event density within a training batch is uneven. The
network has three outputs:
\begin{itemize}
\item \textbf{Lip heatmap} $P = \sigma(z)$, from the first decoder channel, the lip logit $z$. It is the deployed output: thresholded and reduced to a single
lip region at inference.
\item \textbf{Uncertainty map} $s$, from the second decoder channel, a per-pixel log-variance on the logit. It is supervised by the teacher's confidence $c_t$, so it marks where the teachers disagree, mostly along the mouth boundary, and gives a confidence map alongside the heatmap.

\item \textbf{Auxiliary map} $A$, predicted directly from the hidden state by a single $1 \times 1$ convolution. It provides an error signal that reaches the
temporal stage without passing through the decoder (deep supervision), which stabilises the representation, and is not used at inference.
\end{itemize}

The full model has 29{,}107 parameters.

\subsection{Training Objective}
Let $P$ be the predicted heatmap, $T$ the pooled target, and $A$ the auxiliary
map. Table~\ref{tab:losses} lists the loss terms. The squared-error terms use
the spatial weight
\begin{equation}
w = \alpha_p\,\mathbf{1}[T > 0.3] + \max(0.3,\, c_t)\,\mathbf{1}[T \leq 0.3],
\label{eq:weight}
\end{equation}
with $\alpha_p = 5$: lip cells keep their full weight, while background cells
are discounted where the teacher is uncertain, so an uncertain lip edge lowers
the cost of painting it but never the cost of missing it. The target is raised
to $T^{\gamma}$, $\gamma = 0.75$, which lifts its soft edges.

\textbf{Uncertainty.} Following~\cite{kendall2017uncertainties}, $s$ is the
variance of Gaussian noise on the logit, giving in closed form $k^2 = 1 + \tfrac{\pi}{8}e^{s}$ and an attenuated prediction $\tilde{P} = \sigma(z/k)$, which the squared-error term compares with the target. The predicted confidence $\hat{c} = 1/k^2$ is trained against $c_t$ with a class-balanced cross-entropy; the attenuation uses $k$ without gradient, so $s$ is learned from this term alone and marks where the teachers disagree.

The Tversky term weighs missed lip mass above extra mass and, with the focal term, prevents the collapse to an empty map. The false-positive term makes response far from the mouth cost more than response just outside it. The suppression term acts on windows whose target is effectively empty. The displacement term rewards the predicted lip centre moving as much as the target centre between consecutive windows.

\begin{equation}
\begin{aligned}
L = {}& \lambda_{\mathrm{mse}} L_{\mathrm{mse}}
      + \lambda_{\mathrm{tv}} L_{\mathrm{tv}}
      + \lambda_{\mathrm{foc}} L_{\mathrm{foc}}
      + \lambda_{\mathrm{fp}}  L_{\mathrm{fp}} \\
    & + \lambda_{\mathrm{sup}} L_{\mathrm{sup}}
      + \lambda_{\sigma} L_{\sigma}
      + \lambda_{\mathrm{disp}} L_{\mathrm{disp}}
      + \lambda_{\mathrm{deep}} L_{\mathrm{deep}},
\end{aligned}
\label{eq:loss}
\end{equation}
where the terms $L_{\mathrm{mse}}$ to $L_{\sigma}$ of each window are weighted by a difficulty score derived from the teacher labels and normalised by its mean over the training set: the share of the window's events labelled non-lip, each weighted by its proximity to the teacher's lip centre and up-weighted below the lip line.

\begin{table*}[t]
  \centering\small

\begin{tabular}{@{}lrcccc|c|ccccc|ccc@{}}
\toprule
& & \multicolumn{5}{c|}{On lips (\%) $\uparrow$} & Med. & P90 & $\leq$10\,px & IoU$\geq$0.5 & Eye & Switch & Jitter & Streak \\
Model & Params & S10 & S11 & S12 & S13 & All & px $\downarrow$ & px $\downarrow$ & \% $\uparrow$ & \% $\uparrow$ & \% $\downarrow$ & /s $\downarrow$ & cells $\downarrow$ & ms $\uparrow$ \\
\midrule
Ours & \textbf{29{,}107} & 91.7 & 75.7 & 31.9 & \textbf{79.3} & 69.5 & 16.2 & 144 & 24.6 & 35.9 & 6.2 & \textbf{2.1} & 2.8 & \textbf{326} \\
CNN & 164{,}650 & \textbf{96.9} & 82.7 & 47.6 & 74.5 & \textbf{75.6} & \textbf{11.2} & \textbf{81} & \textbf{44.9} & \textbf{43.2} & \textbf{3.9} & 8.9 & 6.0 & 89 \\
U-Net & 165{,}846 & 87.5 & \textbf{91.5} & 40.1 & 31.9 & 62.2 & 13.9 & 140 & 40.3 & 43.1 & 15.7 & 11.0 & 6.6 & 63 \\
Ours + deep encoder & 50{,}221 & 89.5 & 90.9 & \textbf{65.3} & 17.6 & 66.4 & 13.8 & 115 & 31.1 & 22.0 & 8.6 & \textbf{2.1} & \textbf{1.9} & 281 \\
\bottomrule
\end{tabular}
  \caption{Unseen simulated speakers S10--S13}
  \label{tab:main}
\end{table*}

\begin{table}[t]
\centering
\small
\setlength{\tabcolsep}{4pt}
\caption{Loss terms. $\tilde{T}$ is the target dilated by a $7 \times 7$ max pool, $T_G$ the target max-pooled to grid resolution, $w_G$ the weight $w$ on the grid with threshold $0.15$, and $c(\cdot)$ the soft-argmax centre of a map thresholded at $0.3$.}
\label{tab:losses}
\begin{tabular}{@{}llc@{}}
\toprule
Term & Definition & $\lambda$ \\
\midrule
$L_{\mathrm{mse}}$ & $\mathrm{mean}\big(w\,(\tilde{P} - T^{\gamma})^2\big)$ & 1.0 \\
$L_{\mathrm{tv}}$ & Tversky, $\beta = 0.6$ on missed mass & 1.0 \\
$L_{\mathrm{foc}}$ & focal BCE, $\gamma = 2$, $\alpha = 0.75$ & 0.2 \\
$L_{\mathrm{fp}}$ & $\mathrm{mean}\big(P\,\mathbf{1}[T < 0.1]\,(1 - \tilde{T})\big)$ & 4.0 \\
$L_{\mathrm{sup}}$ & $\mathbf{1}[\mathrm{mean}(T) < 0.02]\;\mathrm{mean}(P)$ & 0.5 \\
$L_{\sigma}$ & balanced BCE between $\hat{c}$ and $c_t$ & 0.5 \\
$L_{\mathrm{disp}}$ & SmoothL1 between $\Delta c(P)$ and $\Delta c(T)$ & 0.05 \\
$L_{\mathrm{deep}}$ & $\mathrm{mean}\big(w_G (A - T_G)^2\big) + 0.5\,L_{\mathrm{iou}}(A, T_G)$ & 0.2 \\
\bottomrule
\end{tabular}
\end{table}

\textbf{Training.} Each training clip starts with 64 burn-in windows (0.4\,s)
that only advance the slow state, without gradient, followed by 16 windows
(100\,ms) trained with truncated backpropagation through time (BPTT).

Optimisation uses AdamW with learning rate $10^{-3}$, halved every 4 epochs after
epoch 8, weight decay $10^{-3}$, gradient clipping at 1.0 and an exponential
moving average of the weights (0.999). Half of the clips receive a random smooth head trajectory (translation up to 6\% of the frame, scale $\pm 10\%$) and, on 30\% of their windows, one or two jittered copies of the non-lip events as distractors. The checkpoint is selected on the validation speakers by IoU $\times$ detection rate $\times$ $(1 -$ false-positive rate$)$, and training stops once the learning rate has reached its floor ($5\times10^{-5}$) without improvement for 6 epochs.

\textbf{Inference.} The heatmap is thresholded at $0.3$ and split into connected components; no output is given when the heatmap peak is below $0.35$. A lightweight tracker selects the component with the highest peak, rejects jumps of more than 10 cells while a nearer candidate exists, holds the last region for up to 12 windows when nothing is detected, and re-anchors when a better component persists; the output is the heatmap-weighted centroid of the selected component. The same tracker is applied to every compared model.

\subsection{Baselines}
\label{sec:baselines}
Three further models are implemented on the same polarity-separated voxel
input and heatmap output for comparison.
\begin{itemize}
\item \textbf{CNN:} a fully convolutional encoder--decoder, downsampling from $64 \times 64$ to $8 \times 8$ and back, each
level a pair of $3 \times 3$ convolutions with group normalisation and GELU
(164{,}650 parameters).
\item \textbf{U-Net}~\cite{ronneberger2015u}: the same levels with skip
connections between encoder and decoder (165{,}846 parameters).
\item \textbf{Ours + deep encoder:} the proposed model with a deep spatial
encoder, a small U-Net with skip connections over the window's event image
whose output is added to every bin's encoding, and a high-resolution skip path
that corrects the decoder output at $64 \times 64$ (50{,}221 parameters). 
\end{itemize}

 The CNN and U-Net are trained with MediaPipe labels instead of the supervision target, without deep supervision and with a standard heatmap loss; the heatmap target is only assigned where events exist in the label.

\section{Experimental Results}

\textbf{Protocol.} Models are evaluated on four unseen speakers S10--S13, 59{,}846 windows in total; windows with interpolation errors or no detectable face are excluded. A prediction is \emph{on lips} when its centre falls in the lip box, and on the \emph{eye} when its centre is closer to the eye than to the lip. 

During development, early models included too many sparse events from facial features (like the beard) in the predicted region. As the predicted region became more precise and smaller, the errors moved to the eye, which has a similar event pattern to the lip; the eye rate is therefore reported.

Precision is reported as the median and 90th-percentile centre error, the share of windows within 10\,px, and the share with IoU $\geq 0.5$. Stability is reported as \emph{switches}, the number of times per speaking second that the output leaves the lips for another location, the mean \emph{streak}, the time the output stays on the lips without interruption, and \emph{jitter}, the RMS window-to-window motion of the predicted centre beyond that of the reference, in cells of the $64\times64$ heatmap. 

\subsection{Baseline comparison}

\begin{figure}[h]
  \centering
  \includegraphics[width=0.7\columnwidth]{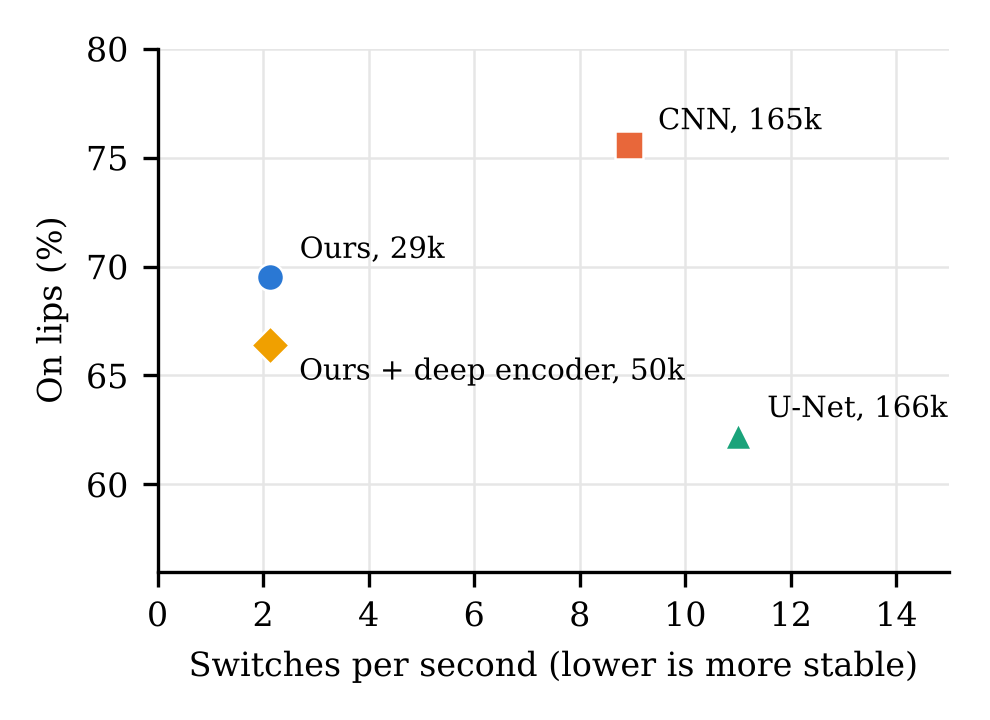}
  \caption{On-lips rate against switches per second.}
  \label{fig:tradeoff}
\end{figure}

Table~\ref{tab:main} compares the baselines trained on the same data. The CNN is the best when it comes to precision, at 75.6\% on lips with a median error of 11.2\,px, and 44.9\% of its windows fall within 10\,px. Second is the proposed model, with 69.5\%, 16.2\,px and 24.6\%. The proposed model is the most stable. It leaves the lips 2.1 times per second against 8.9 for the CNN and 11.0 for the U-Net, and once on the lips it stays there for 326\,ms on average, 3.7 times longer than the CNN. Its jitter is less than half that of the CNN, 2.8 against 6.0 cells. The deep-encoder variant matches this stability, 2.1 switches per second with the lowest jitter of 1.9 cells, at a lower on-lips rate of 66.4\%. The U-Net confuses the lips with the eyes most often, in 15.7\% of windows. Fig.~\ref{fig:tradeoff} shows this trade-off between precision and stability.

\subsection{Scale Shift and Spatial Context Trade-off}
\label{sec:modeleval}

The memory is stress tested by resetting the slow state at every window and by removing each pathway, with the weights unchanged; this experiment is a test-time ablation without retraining. Without the memory, the model becomes more willing to predict, in 99.8\% of windows against 98.6\%, but it is a lot less precise and less stable. Switching rises from 2.1 to 9.3 per second, the mean streak drops from 326 to 67\,ms, close to the CNN and U-Net baselines, and both the eye hits (6.3 to 11.5\%) and the median error (16.2 to 23.3\,px) increase. Removing the slow pathway completely has the same effect and worsens the results further. Removing the fast pathway gives the opposite trade-off: the model becomes more stable, with 0.8 switches per second and half the jitter, but less precise, with a median error of 17.4\,px. The trade-off is that the slow state holds the prediction on the lip, while the fast pathway keeps it responsive and precise, so the design choice of holding both stands. This design, however, has a speed trade-off: the 16 bins per window take about 5\,ms, against 1.4\,ms for the CNN on the same GPU, both still below 6.25\,ms.

\begin{table}[h]
  \centering\small
  \resizebox{\columnwidth}{!}{
\begin{tabular}{@{}lcccccc@{}}
\toprule
Setting & Med. px $\downarrow$ & Eye \% $\downarrow$ & Detected \% $\uparrow$ & Switch /s $\downarrow$ & Jitter cells $\downarrow$ & Streak ms $\uparrow$ \\
\midrule
Full model & 16.2 & 6.3 & 98.6 & 2.14 & 2.78 & 326 \\
w/o fast pathway & 17.4 & 5.7 & 95.8 & 0.78 & 1.37 & 712 \\
w/o slow pathway & 25.1 & 11.6 & 99.7 & 10.14 & 7.90 & 59 \\
w/o cross-window state & 23.3 & 11.5 & 99.8 & 9.30 & 7.40 & 67 \\
\bottomrule
\end{tabular}}
  \caption{Test-time ablation of the proposed model.}
  \label{tab:ablation}
\end{table}

Table~\ref{tab:main} shows that the deep-encoder variant performs a lot better on S12, which was a common outlier for all baselines (the U-Net failed on both outliers); the deep-encoder variant got 65.3\% on lip, but it creates a new outlier in S13, with 17.6\% compared to 79.3\%. S13 is an interesting speaker: the camera was very close to them, so their features appear at a different scale. If the memory generalises, why does S13 drop when a deep encoder is added?
\begin{figure}[h]
  \centering
  \includegraphics[width=0.8\columnwidth]{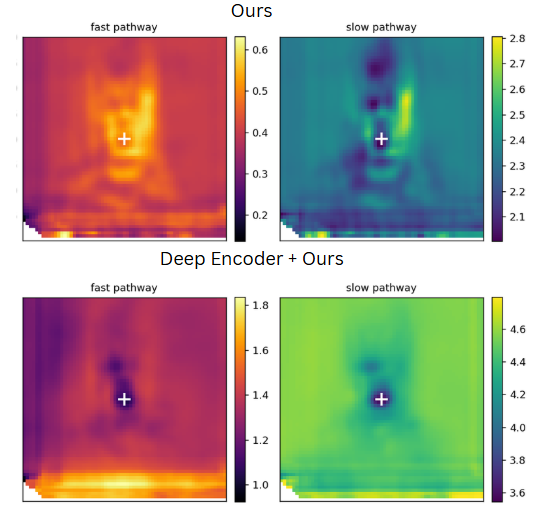}
  \caption{Fast and slow pathway states of the proposed model and of the deep-encoder variant.}
  \label{fig:pathways}
\end{figure}

Looking at the feature maps of the pathways in Fig.~\ref{fig:pathways}, the memory no longer sees most of the face features by the time it receives the input from the encoder. The lip target is already isolated, with no reference point in the face, which stops the model from generalizing to a new scale. This also explains why CNN models are generally more precise: they fit themselves closely to the supervised target, but cannot learn the general features of the face. This was a common issue as well with the U-Net, which dropped to 31.9\% on S13; removing the skip connections from the deep-encoder variant did not help, and it performed similarly.

\subsection{Failure Analysis}
\label{sec:failure}
This section showcases where the proposed model fails, visualising the limitations that were discussed earlier and discussing test cases and their failure points. By now the deep-encoder variant is established to be more precise and has better jitter compared to the proposed model, but as mentioned before, changing the scale of features makes it fail.
\begin{figure}[h]
    \centering
    \includegraphics[width=1\columnwidth]{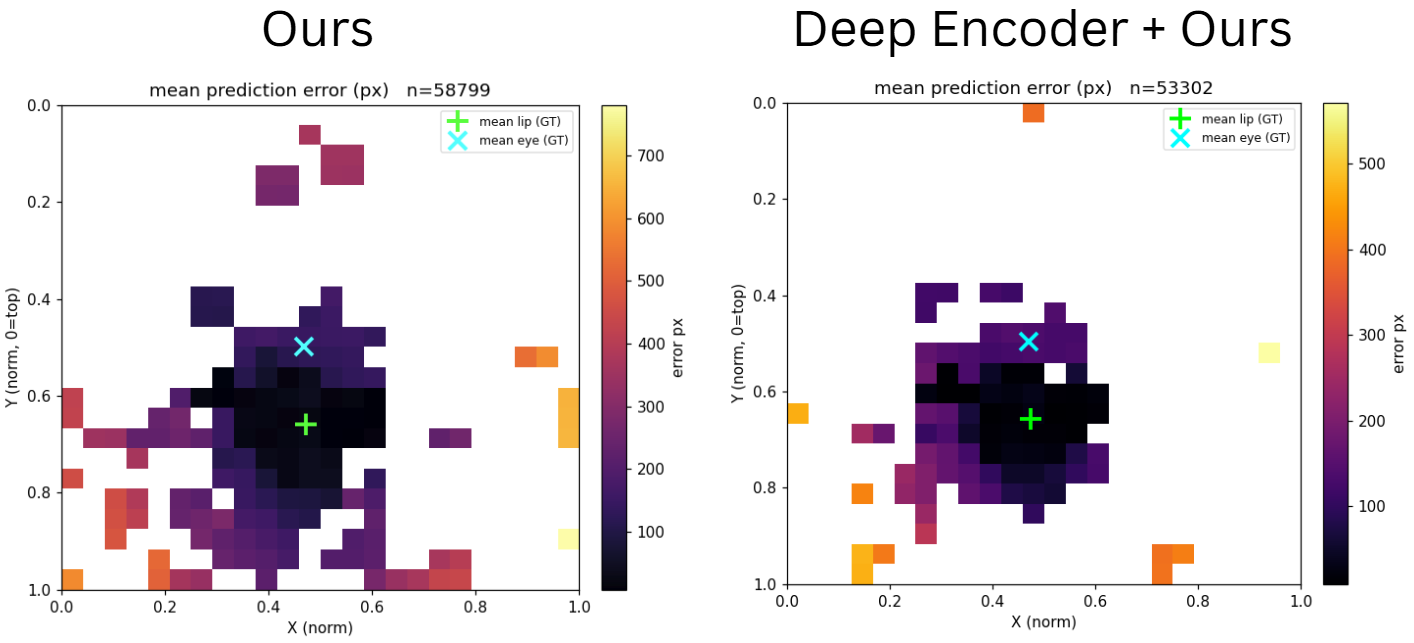}
    \caption{Error maps of the proposed models.}
    \label{fig:errmap}
\end{figure}

\begin{figure}[h]
    \centering
    \includegraphics[width=0.8\columnwidth]{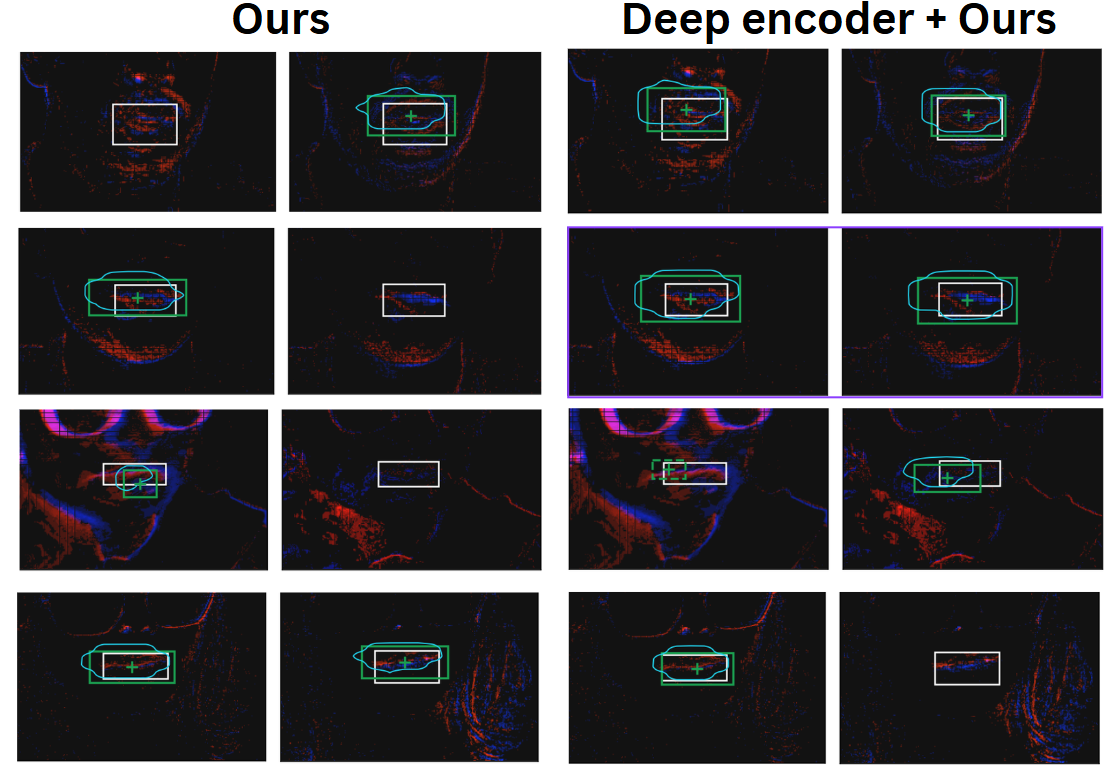}
    \caption{ Example predictions \\ Note: cropped to preserve the anonymity of the volunteers }
    \label{fig:lip_map_track}
\end{figure}

The deep-encoder variant failed in that test case because its predictions shifted down (Fig.~\ref{fig:lip_map_track}), not because it is less precise; the features that help it anchor the results on a new scale are discarded at the encoder level, but with other speakers it is better. Fig.~\ref{fig:errmap} shows the error maps of both models.
Establishing this weakness of the proposed model can only be done through an outlier removal procedure between the variants: removing the proposed model's outlier and the variant's outlier, the proposed model is weaker even if a full comparison shows it is better; on the remaining sequences of S10 and S11, its hit rate is 91.6\% and the variant's is 95.8\%, with a median error of 12.0\,px against 9.5\,px.

The design of the test cases takes into account testing what the model was trained on and what it was not. S10 was recorded normally with little head motion, only enough to generate events, and the model is on lip 91.7\%. S11 was recorded in a different setting to the majority of the test data but also had a similar case of small head movement, on lip at 75.7\%. S12 is the hardest speaker for the proposed model and the CNN due to the head motion; the model fails with 31.9\% on lip, with the highest eye hit at 12.3\%. S13 is recorded close to the camera, and the model still hits 79.3\% on lip, with an eye rate of 5.9\%. The deep-encoder variant was created to solve the S12 outlier, but instead it failed in generalizing a close target, with only 17.6\% on lip on S13.

\subsection{Real Event-Sensor Transfer}
The biggest advantage the proposed model gains is in its real sensor results, compared with the CNN, the strongest baseline model. FES~\cite{s24051409} is not a lip-reading dataset, so only a subset of clips where the speakers were speaking was picked. Seen in Table~\ref{tab:fes}, the proposed model outperforms the CNN on FES, with 58.0\% on lips against 43.2\%, a median error of 17.6\,px against 28.4\,px, and far fewer eye hits (2.3\% against 11.9\%). It is also more stable, with 5.0 against 12.4 switches per second and 126\,ms against 42\,ms streaks. Per sequence, the proposed model is better on 8 of the 12 clips.

\begin{table}[t]
    \centering
    \caption{Transfer to a real event sensor (FES). Both models are trained only on simulated data.}
    \label{tab:fes}
    \resizebox{\columnwidth}{!}{%
    \begin{tabular}{@{}l|c|ccccc|ccc@{}}
    \toprule
    & On lips & Med. & P90 & $\leq$10\,px & IoU$\geq$0.5 & Eye & Switch & Jitter & Streak \\
    Model & \% $\uparrow$ & px $\downarrow$ & px $\downarrow$ & \% $\uparrow$ & \% $\uparrow$ & \% $\downarrow$ & /s $\downarrow$ & cells $\downarrow$ & ms $\uparrow$ \\
    \midrule
    Ours & \textbf{58.0} & \textbf{17.6} & \textbf{208} & \textbf{26.0} & \textbf{27.1} & \textbf{2.3} & \textbf{5.0} & \textbf{5.1} & \textbf{126} \\
    CNN  & 43.2 & 28.4 & 414 & 10.6 & 17.8 & 11.9 & 12.4 & 13.9 & 42 \\
    \bottomrule
    \end{tabular}}
\end{table}

\begin{figure}[h]
    \centering
    \includegraphics[width=1\linewidth]{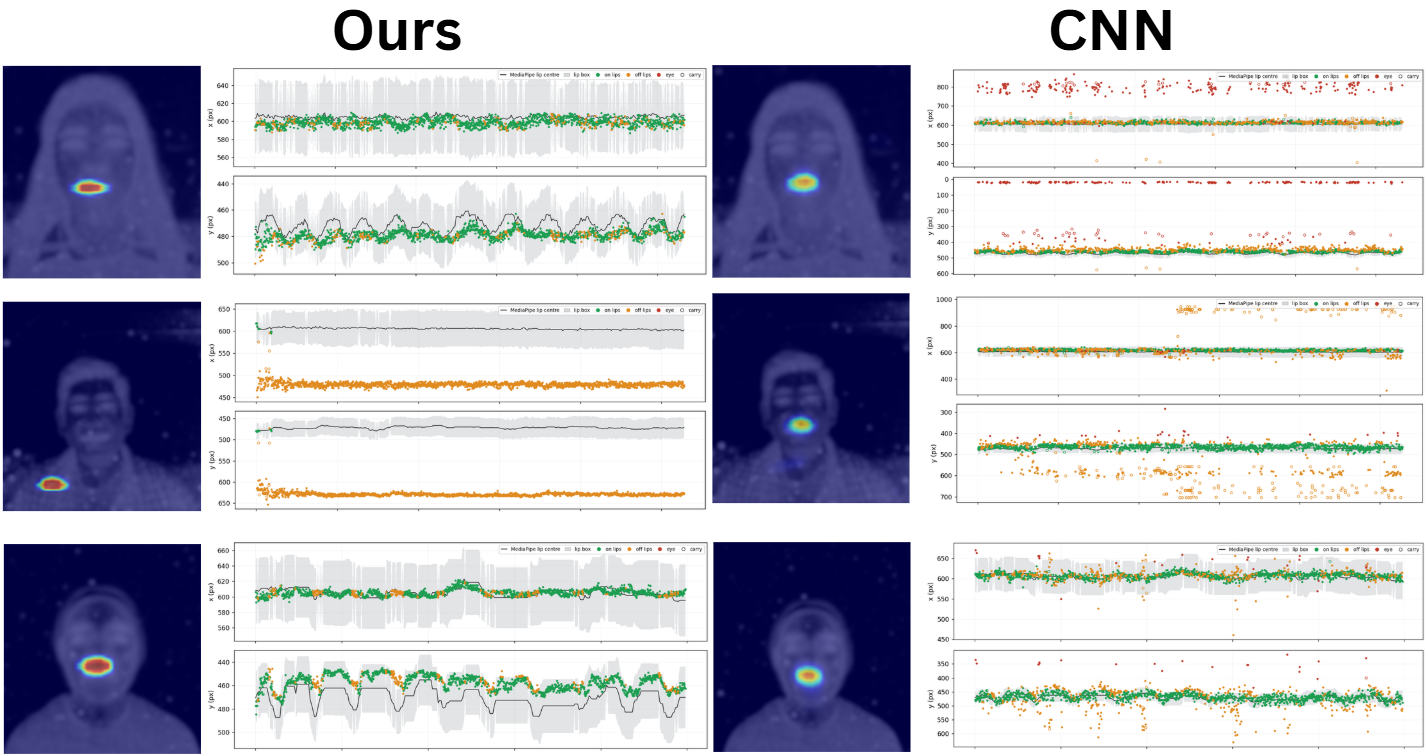}
    \caption{FES examples \cite{s24051409}, predicted center per window against the ground-truth.}
    \label{fig:fes_examples}
\end{figure}

As seen in the top row of Fig.~\ref{fig:fes_examples}, the proposed model is much more stable, with the CNN jumping in the $x$ and $y$ positions. The second row shows a big weakness of the proposed model: when the memory commits to the wrong prediction, it stays there. The model started on the lip, then moved to a wrong region and stayed there, while the CNN, predicting without memory, recovers from wrong predictions. The bottom row shows a case where both find the lip to an equivalent measure, and the proposed model is still much more stable.

\begin{figure}[h]
    \centering
    \includegraphics[width=1\linewidth]{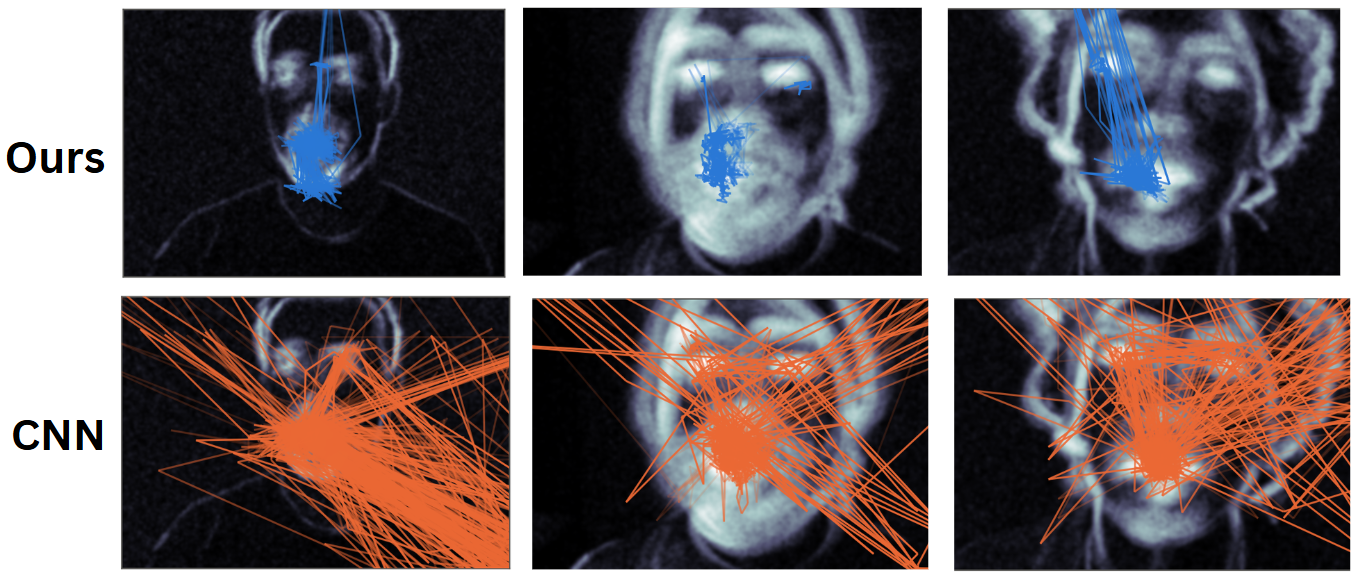}
    \caption{Jitter on FACEMORPHIC \cite{becattini2024neuromorphic2}}
    \label{fig:fm_trails}
\end{figure}

As FACEMORPHIC~\cite{becattini2024neuromorphic2} RGB is not well aligned, there is no ground truth to test; however, stability can be tested, shown in Fig.~\ref{fig:fm_trails} where every predicted centre of a recording is connected to the next one and drawn on a single frame. The proposed model stays in a compact cluster around the mouth, with a few jumps. The CNN jumps to the background often, predicting outside the face during silent activity and anchoring to the noise, while in the recordings inspected the proposed model predicted less during silent behavior.

\section{Discussion}
\label{sec:discussion}
The discussion follows the three gaps of the problem statement, then the transfer to a real sensor.  

\textbf{Task and output:} The proposed model localizes the lip region as a dense map from the full event frame, without a face detector or crop, the input that lip reading, speaker recognition, and voice activity detection consume. It is on the lips in 69.5\% of windows on unseen speakers; however, the CNN outperforms it in precision, with a median error of 11.2 against 16.2\,px.  

\textbf{Representation:} Unlike time surfaces or summed-polarity grids, the polarity-separated voxel grid keeps both event order and polarity, so this gap is addressed by design; the baselines use the same grid and share this. The grid keeps event order in bins of fixed duration, which the fast pathway reads in order within a window; removing it raises the median error from 16.2 to 17.4\,px (Table~\ref{tab:ablation}).

\textbf{Temporal stability:} Temporal stability is the advantage of the proposed model over the CNN: it leaves the lip 2.1 times per second against 8.9, holds an on lip streak 3.7 times longer, and has half its jitter. Resetting the slow state at every window raises switching to 9.3 per second, close to the baselines, linking this stability to the memory. The slow pathway, with $\tau$ up to 2\,s, keeps most of its state across windows and anchors the prediction, while the fast pathway follows the lip motion and keeps the model responsive.

\textbf{Transfer to a real sensor:} On FES, the proposed model leads the CNN on every metric (Table~\ref{tab:fes}), with 58.0\% against 43.2\% on lips and a jitter of 5.1 against 13.9 cells; on FACEMORPHIC, during silent behaviour, the CNN predicts on background noise, while in the inspected recordings the proposed model predicts less (Fig.~\ref{fig:fm_trails}). The real sensor changes the contrast threshold and noise of the events; this does not prove the effectiveness of the mechanism, but it shows that the temporal memory generalizes better to it than the frame-wise CNN.

In summary, the results address the task and temporal stability gaps, and the design addresses the representation gap; the memory brings stability and generalization to the real sensor, at the cost of precision against the CNN on simulated data.

\section{Limitations and Future Work}
\label{sec:limitations}
Training data is limited to simulated events with a fixed contrast threshold, and on the real sensor the on-lips rate drops. The frame-wise CNN outperforms the proposed model in precision; the advantage of the proposed model is temporal stability. The memory that creates this stability is also a weakness: once the recurrent state commits to a wrong region, the model stays there, while the CNN recovers (Fig.~\ref{fig:fes_examples}, second row). S12 shows a weakness under head motion, and the deep-encoder variant recovers S12 but, relying on spatial layout, breaks under a new scale.

Future work could add a recovery mechanism for when the state commits to a wrong region, scale-aware augmentation, and a direct comparison of event representations, and could combine the localized ROI with the pathway memory for event-based lip reading without a manual crop.

\section{Conclusion}
\label{sec:conclusion}
Mouth movements precede the voice and are fused with it within about 200\,ms, a window that event cameras record at a much higher temporal resolution than frame cameras. Event-based lip applications, however, still depend on a lip ROI given by hand or by the dataset, while existing event-based methods localize the mouth only as a point or a set of landmarks; localizing the lip region directly from the events is the front end they are missing, and this work provides it.

The proposed model predicts the lip region from uncropped event streams with a polarity-separated voxel grid and a two-timescale continuous-time memory trained with teacher uncertainty. The frame-wise CNN outperforms it in precision, but the proposed model tracks the mouth more robustly, with about four times fewer switches and longer streaks. On a real event sensor, it outperforms the CNN in both on-lips rate and temporal stability despite training only on simulated events, and on FACEMORPHIC it keeps a compact track while the CNN jumps across the frame. The results show that temporal memory transfers better than spatial appearance for event-based lip localization, and that temporal stability should be reported alongside accuracy.
{
    \small
    \bibliographystyle{ieeenat_fullname}
    \bibliography{main}
}
\end{document}